\documentclass[lettersize,journal]{IEEEtran}
\usepackage{amsmath,amsfonts}
\usepackage{algorithmic}
\usepackage{algorithm}
\usepackage{array}
\usepackage{booktabs}
\usepackage[caption=false,font=normalsize,labelfont=sf,textfont=sf]{subfig}
\usepackage{textcomp}
\usepackage{stfloats}
\usepackage{multirow}
\usepackage{url}
\usepackage{verbatim}
\usepackage{graphicx}
\usepackage{amssymb}
\usepackage{booktabs}
\usepackage{amssymb}
\usepackage{bm}
\usepackage{makecell}
\usepackage[table]{xcolor}
\usepackage[colorlinks,linkcolor=blue]{hyperref}
\usepackage{cite}
\begin{document}

\title{Vision-Based UAV Self-Positioning in Low-Altitude Urban Environments}

\author{
Ming Dai,
Enhui Zheng,
Zhenhua Feng,~\IEEEmembership{Senior~Member,~IEEE},
Lei Qi,
Jiedong Zhuang,
and Wankou Yang,
\thanks{M. Dai, L. Qi and W. Yang are with the School of Automation, Southeast University, Nanjing 210096, China (emails: 869906992@qq.com; wkyang@seu.edu.cn; qilei@seu.edu.cn.}
\thanks{E. Zheng is with the Unmanned System Application Technology Research Institute, China Jiliang University, Hangzhou 310018, China (email: ehzheng@cjlu.edu.cn).}
\thanks{Z. Feng is with the School of Computer Science and Electronic Engineering, University of Surrey, Guildford GU2 7XH, UK (email: z.feng@surrey.ac.uk).}
\thanks{J. Zhuang is with Zhejiang University, Hangzhou 310063, China (email: 312788727@qq.com).}
}

\maketitle

\begin{abstract}
Unmanned Aerial Vehicles (UAVs) rely on satellite systems for stable positioning. However, due to limited satellite coverage or communication disruptions, UAVs may lose signals from satellite-based positioning systems. In such situations, vision-based techniques can serve as an alternative, ensuring the self-positioning capability of UAVs. However, most of the existing datasets are developed for the geo-localization tasks of the objects identified by UAVs, rather than the self-positioning task of UAVs. Furthermore, the current UAV datasets use discrete sampling on synthetic data, such as Google Maps, thereby neglecting the crucial aspects of dense sampling and the uncertainties commonly experienced in real-world scenarios. To address these issues, this paper presents a new dataset, DenseUAV, which is the first publicly available dataset designed for the UAV self-positioning task. DenseUAV adopts dense sampling on UAV images obtained in low-altitude urban settings. In total, over 27K UAV-view and satellite-view images of 14 university campuses are collected and annotated, establishing a new benchmark. In terms of model development, we first verify the superiority of Transformers over CNNs in this task. Then, we incorporate metric learning into representation learning to enhance the discriminative capacity of the model and to lessen the modality discrepancy. Besides, to facilitate joint learning from both perspectives, we propose a mutually supervised learning approach. Last, we enhance the Recall@K metric and introduce a new measurement, SDM@K, to evaluate the performance of a trained model from both the retrieval and localization perspectives simultaneously. As a result, the proposed baseline method achieves a remarkable Recall@1 score of 83.05\% and an SDM@1 score of 86.24\% on DenseUAV.
The dataset and code will be made publicly available on \url{https://github.com/Dmmm1997/DenseUAV}.

\end{abstract}


\section{Introduction}\label{sec1}
\IEEEPARstart{I}{n} recent years, Unmanned Aerial Vehicles (UAVs) have become increasingly vital in various applications, such as agricultural operations, ground reconnaissance, and civilian aerial photography~\cite{bib5,bib7,FSRA,DBLP}. With the advancements in onboard computing capabilities and lightweight algorithms, different visual algorithms like object tracking, object detection, and Simultaneous Localization And Mapping (SLAM)~\cite{drone_objdet,bib39,DOD,TCSVT_tracking} have been widely deployed in UAVs. Notably, recent developments in UAV-related visual localization tasks have emerged. The pioneering University-1652 dataset~\cite{bib5} introduces the use of drone view for cross-view geo-localization. Additionally, University-1652 proposes tasks such as drone-view target localization and drone navigation, which extend the applications of UAV geo-localization. 

However, as the applications of UAVs become more widespread, ensuring the stability of UAV self-positioning has emerged as a crucial concern. Typically, the acquisition of positioning information heavily relies on satellite systems. This approach necessitates the stability of satellite signals and the absence of signal disturbances. In reality, due to the fixed frequency band of GPS signals and the potential for UAVs to operate in environments with obstructions or signal interference, maintaining a stable connection becomes challenging. Once a UAV loses awareness of its position, it can result in a loss of control, underscoring the significance of ensuring reliable self-positioning capabilities for UAVs.

Considering the aforementioned challenges, we propose a novel UAV-based geo-localization task, namely \textit{UAV self-positioning}, to address the need for accurate self-positioning of UAVs in environments without GPS. In contrast to the University-1652 dataset, we focus on determining the position of UAVs rather than the captured objects. This aligns with the goals presented in previous works~\cite{bib3, bib4}. However, our approach differs in that we do not employ panoramic images as the UAV-view image. Instead, we offer a simpler solution by vertically orienting the UAV camera towards the ground, thus mitigating view deviation issues.

To effectively train a model for the UAV self-positioning task, it is necessary to develop a dataset specifically tailored for this purpose. In this paper, we propose a new dataset, DenseUAV, that has several key features.
\begin{itemize}
    \item Real scene sampling: Unlike previous UAV-related geo-localization datasets~\cite{bib5, SUE-200} that rely on synthetic data collected from platforms like Google Maps, DenseUAV captures real scenes. While synthetic data is readily accessible and useful, it exhibits significant disparities with actual scenes due to distinct domains, variations in viewing angles, and temporal offsets.

    \item Dense sampling: Previous datasets primarily focused on mining salient features, such as buildings, overlooking the impact of overlapping regions between adjacent images on positioning. DenseUAV emphasizes dense sampling, enabling high-precision UAV self-positioning.

    \item Multi-heights sampling: DenseUAV incorporates samples captured at three different heights, enhancing the model's robustness to varying UAV flying heights.

    \item Multi-time node satellite images: During the training process, we introduce multiple temporal satellite images to facilitate the model's adaptability to spatial changes caused by temporal offsets.

    \item Multi-scale satellite images: The construction of the satellite imagery in DenseUAV incorporates multi-scale image sets, strengthening the model's resilience to varying scales of satellite images.
\end{itemize}

At the model level, we construct a robust baseline model for UAV self-positioning based on a vision transformer architecture. In addition, we use a Siamese network that allows the model to learn shared representations across different modalities. Extensive investigations are completed to analyze the impact of various components on the UAV self-positioning task. These include data augmentation, backbone networks, prediction heads, and loss functions. 
Secondly, we introduce metric learning and mutual learning to our baseline method. Metric learning is motivated by the UAV self-positioning task involving image matching across different modalities. It addresses the modality discrepancy issue effectively and enhances the model's discriminative power. Moreover, the incorporation of mutual learning facilitates concurrent learning between different modalities, ensuring their alignment throughout the training process. 
\begin{figure*}[!t]
\centering
\includegraphics[width=0.99\linewidth]{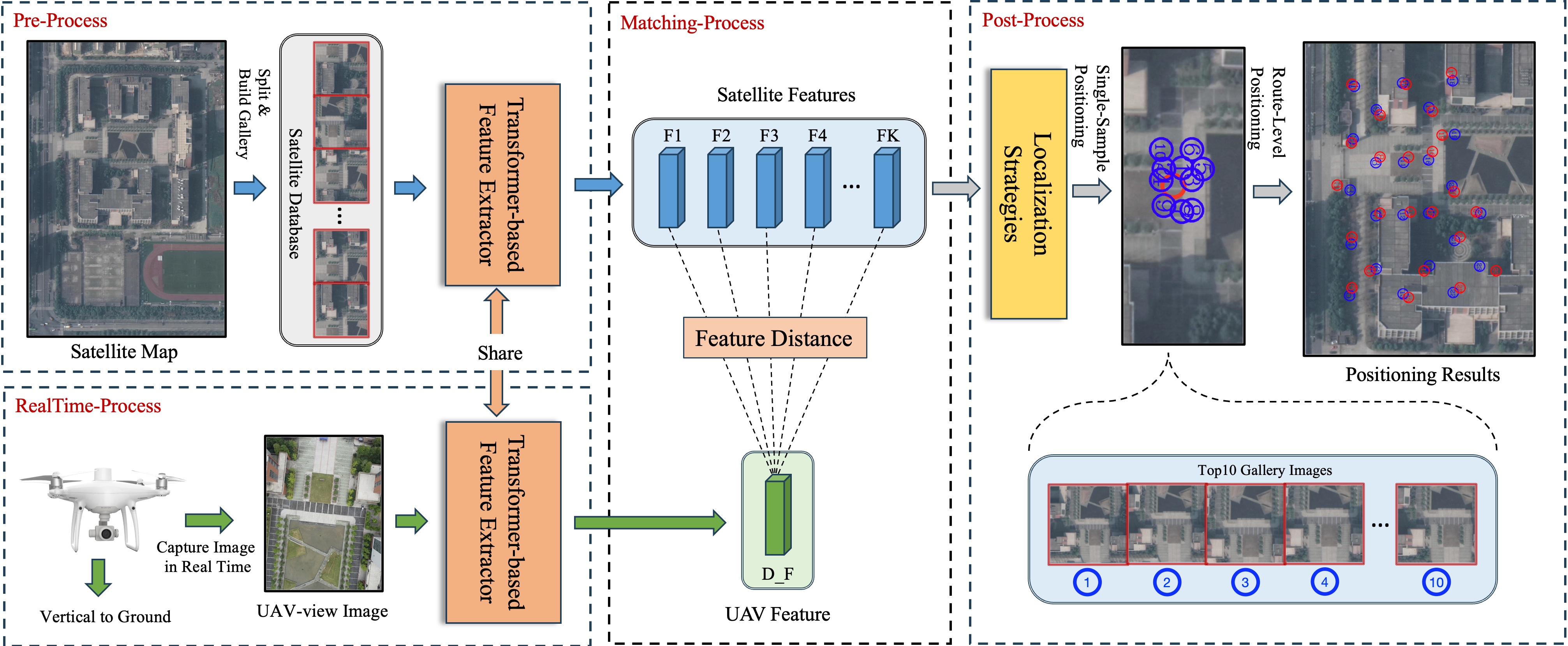}
\caption{The overall flow chart of UAV self-positioning. \textit{Localization Strategies} refer to post-processing operations after model inference, such as Global-Search and Neighbor-Search. Blue circles in \textit{Positioning Results} are ground-truth, red is the predicted location, and the numbers indicate correspondence.}
\label{fig2}
\vspace{-0.3cm}
\end{figure*}

We provide a UAV self-positioning pipeline, as shown in Fig.~\ref{fig2}, that has four main steps. In the pre-processing stage, it is imperative to slice the satellite images, acquired from 20-levels Google Maps and subsequently construct a satellite database. Furthermore, we extract their image features \{F1, F2, ..., FN\} using a vision transformer model. 
In the real-time processing stage, the vertical ground camera of the UAV is employed to capture the corresponding UAV-view image, and the associated image feature D\_F is obtained. Subsequently, in the matching stage, we calculate the feature similarity between the UAV and satellite gallery images, resulting in the generation of a corresponding feature distance matrix. Last, the post-processing stage obtains optimal matching through a positioning strategy, thereby enabling continuous positioning. 

The evaluation metric is another essential element. Currently, Recall@K\cite{bib29,bib30} is the mainstream evaluation metric for image retrieval tasks. 
This evaluation indicator is discrete, either 1 or 0. However, the task of UAV self-positioning is spatially continuous. As shown in Fig.~\ref{fig_evaluation}-(a), if the UAV image obtains positive samples according to the precise location, only satellite images completely corresponding to their positions can be used as positive samples (red dotted box). The rest of the satellite images are all regarded as negative samples (blue dashed box), although they have only a small spatial error. Evaluating the self-positioning task with retrieval indicators is obviously not suitable, but the use of L1 and L2 distances can not fully reflect the effectiveness of retrieval. Therefore, we proposed SDM@K to consider both the retrieval and localization performance.

In summary, the main contributions of this paper include:
\begin{itemize}
\item {We construct a low-altitude urban scene dataset, DenseUAV, which is constructed by dense sampling from the UAV's down-looking camera in real scenarios.}
\item {We propose the UAV self-positioning task and establish a full pipeline for the task. 
We also develop a baseline model using multi-task learning and vision Transformer.}
\item {We propose a new evaluation metric, SDM@K, which considers the accuracy of UAV self-positioning from both perspectives of retrieval and localization.}
\item {We extensively evaluate the proposed method and the impact of some key components on its performance.}
\end{itemize}

The rest of the paper is organized as follows. We first introduce the related work in Section~\ref{sec2}. Then, we present the proposed dataset in Section~\ref{sec3}, and a new evaluation metric is designed in Section~\ref{sec4}. Next, the model architecture is introduced in Section~\ref{sec5}. Last, the experimental results and visualization are reported in Section~\ref{sec6} and the conclusion is drawn in Section~\ref{sec7}.

\section{Related Work}
\label{sec2}
In this section, we introduce the existing studies by dividing them into two categories: \textit{geo-localization datasets} and \textit{deep-learning-based geo-localization methods}.

\subsection{Geo-Localization Datasets}\label{sec2.1}

\subsubsection{Ground-to-Aerial Matching}
\label{sec2.1.1}
The geo-localization task was initially proposed to solve the ground-to-aerial matching problem. Some pioneering studies of \cite{bib1, bib2, bib31} were proposed to use publicly available resources to build image pairs for ground and aerial view images. Later, CVUSA\cite{bib3} constructed image pairs from ground-based panoramic images and satellite images, while CVACT\cite{bib4} added spatial factors to CVUSA, i.e., orientation maps. Recently, VIGOR\cite{VIGOR_2021} changed the previous center-to-center image matching and redefined the problem with a more realistic assumption that the query image can be arbitrary in the area of interest. 

\subsubsection{Drone-to-Satellite Matching}\label{sec2.1.2}
With the development of UAVs, the geo-localization task goes beyond ground-based operations. 
University-1652\cite{bib5} introduced the drone-view into cross-view geo-localization, using the drone-view as a transition view to reduce the difficulty of matching between ground and satellite views, and regarded it as a retrieval task. In addition, University-1652 proposed two UAV-based subtasks, namely drone-view target localization and drone navigation.
Further, SUE-200\cite{SUE-200} improved the adaptability of a model in terms of flight altitudes by collecting drone images at four different altitudes. 

The proposal of DenseUAV was originally inspired by University-1652. However, they have significant differences. University-1652 primarily focused on identifying and localizing buildings with prominent features, and these buildings were discretely distributed in space. In contrast, DenseUAV was designed to address the problem of self-positioning of UAVs in GPS-denied environments. This requires a model to not only consider prominent features but also be aware of the spatial distribution of objects in images.

\begin{table*}[h]
\centering
\caption{A summary of the existing geo-localization datasets.}
\label{table1}
\tiny
\renewcommand\arraystretch{1.2}
\resizebox{1.0\hsize}{!}{
    \begin{tabular}{ccccccccc}
        \specialrule{0.75pt}{0pt}{1pt}
        {Datasets}& 
        {DenseUAV(ours)} &
        {SUES-200\cite{SUE-200}} &
        {University-1652\cite{bib5}}& 
        {VIGOR\cite{VIGOR_2021}}&
        {CVUSA\cite{bib3}} & 
        {Tian et al.\cite{bib2}}
        \\
        \specialrule{0.5pt}{1pt}{1pt}
        
        {Training} & 
        {10$\times$225.6$\times$9}& 
        {120$\times$51} &
        {701$\times$71.64}& 
        {91k + 53k}&
        {35.5k$\times$2} & 
        {15.7k$\times$2}
        \\
        \specialrule{0pt}{0pt}{0pt}
        {Platform} & 
        {Drone, Satellite}&
        {Drone, Satellite}&
        {Drone, Ground, Satellite}&
        {Ground, Aerial}&
        {Ground, Satellite} & 
        {Ground, Aerial} 
        \\
        \specialrule{0pt}{0pt}{0pt}
        {Imgs./Platform}
        & 
        {3 + 6} & 
        {50 + 1}&
        {54 + 16.64 + 1} & 
        {/}&
        {1 + 1}& 
        {1 + 1}
        \\
        \specialrule{0pt}{0pt}{0pt}
        {Target}& 
        {UAV}& 
        {Diverse}&
        {Building}& 
        {User}&
        {User}& 
        {Building}
        \\
        \specialrule{0pt}{0pt}{0pt}
        {Sampling}& 
        {Dense}& 
        {Discrete}&
        {Discrete}& 
        {Discrete}&
        {Discrete}& 
        {Discrete}
        \\
        \specialrule{0pt}{0pt}{0pt}
        {Source}& 
        {Real Scenes}& 
        {Google Map}&
        {Google Map}& 
        {Google Map}&
        {Google Map}& 
        {Google Map}
        \\
        \specialrule{0pt}{0pt}{0pt}
        {Evaluation}&
        {R@K \& SDM@K}& 
        \makecell[c]{R@K \& AP \& RB \& PF} &
        {R@K \& AP}& 
        {Meter Accuary}&
        {R@K}& 
        {PR \& AP}
        \\
        \specialrule{0.75pt}{1pt}{0pt}
    \end{tabular}
}
\end{table*}

\subsection{Deep-Learning-Based Geo-Localization}
\label{sec2.2}

\subsubsection{Supervision-Based Methods}
This category primarily focuses on different types of supervision learning and enhances the discriminative capability of a model by introducing new loss functions. The commonly used methods are based on CNNs supervised by the cross-entropy loss\cite{celoss}, which involved image feature embedding via classification supervision. 
In order to improve the matching reliability, triplet loss\cite{bib6} was utilized to reduce the distance between positive pairs and increase the distance between negative pairs. Additionally, several methods based on triplet loss were proposed to further strengthen the discriminative ability of features\cite{quadruplet, improved, bib10, SEH}. By pulling the distance between positive pairs, contrastive loss\cite{bib8, simclr} could further enhance the performance of a geo-localization model. Moreover, some multi-task supervised learning methods\cite{bib9, bib11, bib12} were introduced subsequently to further boost the discriminative ability of the model.

\subsubsection{Matching-Based Methods}
This category focuses on the spatial information asymmetry between different domains. 
Some existing approaches aim to reduce the variance of different domains through view transformation. 
For example, \textit{Shi et al.} used the polar coordinate transformation\cite{bib17, bib32} to map ground panorama images to a top-down perspective. 
Although this method is highly efficient, it suffers from distortion and information loss. 
For another example, \textit{Toker et al.} used GAN to generate images from the target domain\cite{bib33}. Another focus in this category is to overcome the viewpoint bias and feature misalignment issues between different domains\cite{bib6, sun2019geocapsnet}. 
Some methods have tried to minimize domain differences by utilizing the results of pixel-level segmentation\cite{bib14, bib11}. Also, feature alignment has been used for specific scenes by employing feature chunking. For instance, Local Pattern Network (LPN)\cite{bib18} proposed a square chunking strategy to extract useful information from the edges and the chunking scheme, resulting in significant performance boosting. FSRA\cite{bib28} automatically divides the corresponding instances through the distribution of thermal values and performs metric learning on the corresponding regions of the two domains.
	
\section{The DenseUAV Dataset}
\label{sec3}
In this part, we first introduce the key features of the proposed DenseUAV dataset and emphasize its distinctions from the existing datasets. Then we present the sampling method and the composition of the dataset in detail.

\subsection{Characteristics}\label{sec3.1}
First, it should be highlighted that the aim of the proposed DenseUAV dataset is UAV self-positioning, which has not been addressed in previous datasets. 
Additionally, DenseUAV is a dataset designed for low-altitude urban scenes, comprising perspectives from UAV-view and Satellite-view. 
More details of DenseUAV are shown in Table~\ref{table1}. 
These details encompass a comprehensive analysis of various aspects, such as the quantity of training data, the data composition platform, the specific number of images encompassed within each platform, the designated positioning target, the sampling method employed, the source of the data, and the associated evaluation indicators.
Some notable characteristics of DenseUAV at the data level will be further introduced below.

\subsubsection{Dense Sampling}
\label{sec3.1.1}
Existing public datasets\cite{bib5,bib4} usually use discrete sampling methods based on landmark buildings during the dataset construction phase to capture intricate architectural features. 
Nevertheless, this approach introduces a noticeable disparity between various categories due to the discrete sampling-based library construction. 
Consequently, the retrieval performance during testing is hindered, as the challenge of accurately retrieving the desired information is not sufficiently pronounced. The proposed DenseUAV dataset differs from previous geo-localization datasets mainly in terms of `dense sampling' which refers to the fact that there will be some overlapping areas between adjacent frames. 
In the training process, a model is required to not only capture fine-grained features but also recognize spatial relative information. 
Similarly, interference from neighboring frames significantly affects the testing process thus making the benchmark more challenging.

\subsubsection{Real Scene}
\label{sec3.1.2}
Given the high costs of collecting real-world data, the majority of existing datasets\cite{bib5,SUE-200} used in the research community are obtained from diverse perspectives using Google Map. Another notable distinction of the DenseUAV dataset is that all the UAV-view images are captured from authentic real-world scenes. The rationale behind this lies in the fact that, in real-world applications, the time gap between the UAV-view image and satellite-view image leads to varying degrees of discrepancy at the same location, posing a significant challenge. Moreover, real data is deemed more suitable for practical scenarios, thereby minimizing the disparities between the dataset and real-world landing scenes.

\subsubsection{Others}\label{sec3.1.3}
In addition to the aforementioned features, DenseUAV incorporates three different altitude levels for image sampling, which means that the captured UAV images possess varying spatial coverage ranges. Similarly, during the construction of satellite imagery, we also employ three different scales of images. Moreover, to address the spatial information variations arising from temporal disparities, satellite images are generated by using multiple temporal instances.

\subsection{Dataset Construction}\label{sec3.2}
After discussing the main characteristics of the dataset, this subsection covers the data collection scheme and composition.

\begin{figure*}[!t]%
    \centering
    \includegraphics[width=.9\textwidth]{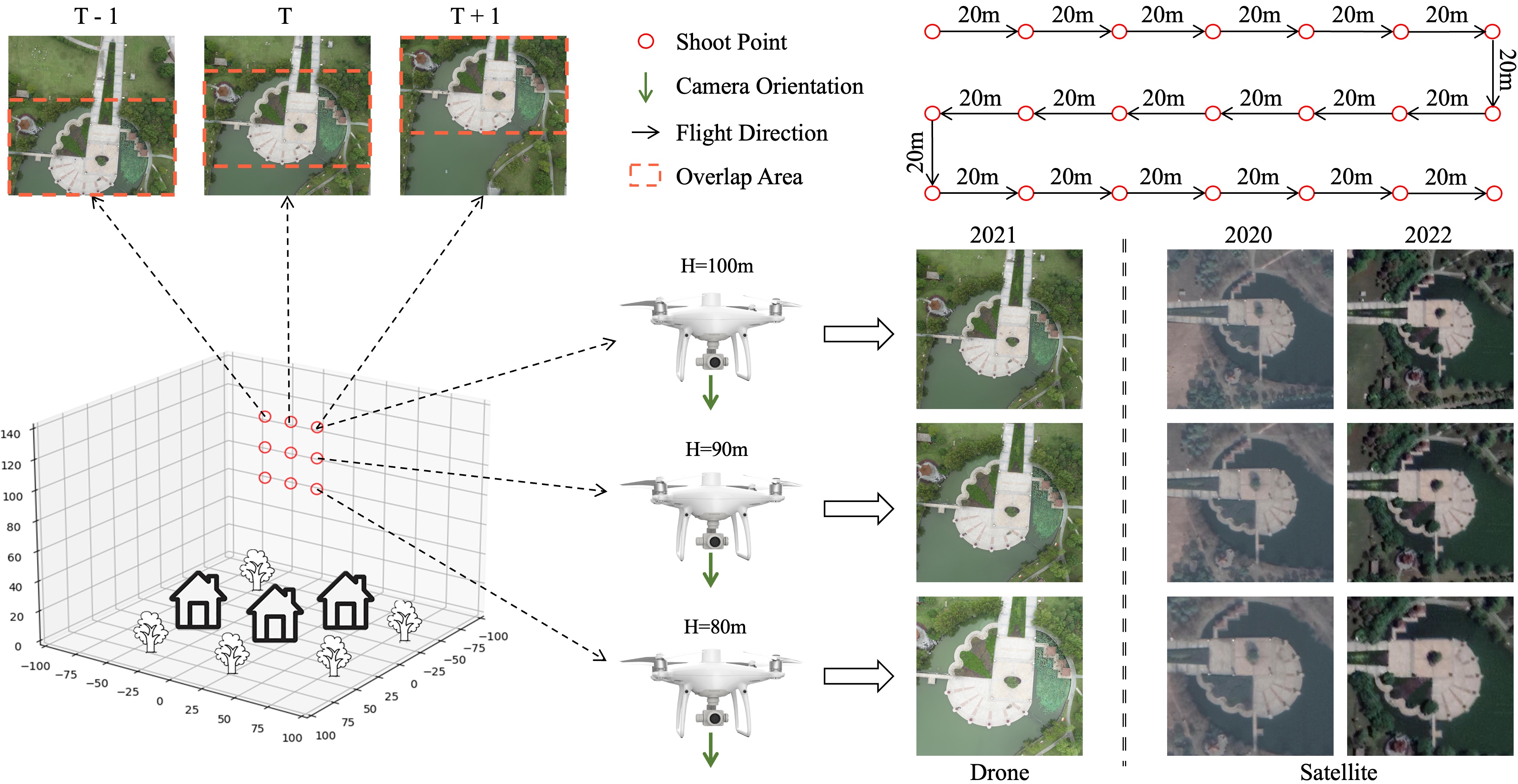}
    \caption{A diagram describing the process of dataset creation. The red circles represent the sampling locations, with an interval of 20 meters, and sampled at 80, 90, and 100 meters altitude, respectively. The green arrows represent the orientation of the camera. T is the current sampling point. T-1 and T+1 are the previous and next sampling points. The orange dotted line represents the overlap area. The upper right corner is the overlooking route, and the black arrow represents the flight direction.}
    \label{fig_dataset}
\end{figure*}

\subsubsection{Sampling Scheme for UAV-View Images}\label{sec3.2.1}
The proposed DenseUAV dataset considers three main factors to capture UAV images. Firstly, by considering the influence of UAV flight height on image scale, the UAV images are collected at three different heights (80m, 90m, and 100m relative to the ground), ensuring nearly identical latitude and longitude coordinates. 
This is achieved through the use of DJI Pilot's waypoint action, with an error control within 1m. 
Secondly, to address the impact of weather and light levels on image quality, a random weather and random time period sampling method is used. 
This includes varying weather conditions (sunny and cloudy days) and random sampling between 6:00 am and 6:00 pm. 
Thirdly, to mitigate perspective bias resulting from changes in UAV camera orientation, a standardized configuration is implemented with the camera facing vertically downward. 
Additionally, rotational data enhancement techniques are employed to address the UAV flight orientation issue (see Section \ref{sec6.3}). 
Furthermore, we set a fixed sampling distance of 20m.
This serves on both the creation of a uniformly spaced dataset and the enhancement of a model's ability to learn subtle differences between classes. 
The upper part of Fig.~\ref{fig_dataset} demonstrates the images captured at adjacent sampling points (T-1, T, and T+1), providing a temporal sequence of images. Meanwhile, the upper right corner of the figure illustrates the flight path of the UAV, which overlooks the scene during data collection.

\subsubsection{Acquisition of Satellite-View Images}
\label{sec3.2.2}
For satellite images, we collect Google Map images at level 20. To enrich the diversity of geographic information and facilitate a model's adaptation to spatial changes over time, two years (2020 and 2022) of satellite images are included in the DenseUAV dataset. Furthermore, to improve the model's resilience to the variations in satellite image scales, three different scales of satellite images are included in the dataset. The visualization results demonstrating different scales are presented in the right part of Fig.~\ref{fig_dataset}.

\subsubsection{Dataset Composition}\label{sec3.2.3}
Regarding the composition of the dataset, we collect real data through UAV sampling at fourteen universities in Zhejiang, China. As presented in Table \ref{table2}, the training set consists of 6768 UAV-view images captured from 2256 sampling points across ten universities and 13536 satellite-view images. Moreover, the test set consists of a query set, including 2331 UAV-view images from 777 sampling points across four universities, and 4662 satellite-view images.
Additionally, the gallery set encompasses a total of 27297 images from 3033 sampling points, covering all fourteen universities.

\begin{table}[!t]
    \tiny
    \centering
    \renewcommand\arraystretch{1.2}
    \caption{The data volume composition of DenseUAV.}
    \label{table2}
    \resizebox{1.0\hsize}{!}{
        \begin{tabular}{ccccc}
            \specialrule{0.75pt}{0pt}{1pt}
            \multirow{2}{*}{Subset} & \multicolumn{2}{c}{\#Imgs} &
            \multirow{2}{*}{\#Classes} & \multirow{2}{*}{\#Universities} \\ \cline{2-3}
            &{UAV-view} &{Satellite-view}& & \\
            \specialrule{0.5pt}{1pt}{1pt}   
            Training & 6768 & 13536 & 2256 & 10 \\
            Query & {2331} & {4662} & 777 & 4 \\
            Gallery & {9099} & {18198} & 3033 & 14 \\
            \specialrule{0.75pt}{1pt}{0pt}
        \end{tabular}
		}
\end{table}

\section{Evaluation Metrics}
\label{sec4}
In this section, we first introduce the widely used evaluation metric, \textit{i.e.,} Recall@K, in cross-view geo-localization. Then we present the proposed SDM@K metric, which consider both retrieval and positioning performance.

\subsection{Spatial Discrete Index (Recall@K)}
\label{4.1}
In image retrieval, Recall@K (R@K) \cite{bib29,bib30} is the most commonly used evaluation metric. Taking R@1 as an example, whether a sample is a correct match can be expressed as:
\begin{equation}
\label{eq1}
I(l_q,l_i)=\left\{\begin{array}{l}1\ \ \ \ \ if\ \
    l_q=l_{i}\  \\
    0\ \ \ \ \ if\ \ l_q\not=l_i\end{array}\right.\ \ \ 
\end{equation}
where $l_q$ corresponds to the category of query, and $l_i$ is the category corresponding to the \textit{i-th} image sorted in the ascending order of the calculated euclidean distance. The resulting value is 1 if it belongs to the same category and 0 if it is not. 
For all the samples, R@1 is defined as:
\begin{equation}
	\label{eq2}
	{\rm R@1}=\ \frac{1}{\left\Vert{}Q\right\Vert{}}\sum_{q\in{}Q}I(l_q,l_1)
\end{equation}
where $Q$ is the set of all query images and $\Vert{}Q\Vert{}$ denotes the number of images in $Q$. 
We can see that the value of R@1 increases when only the category of query and the category of the closest image in the gallery are the same, otherwise, they are regarded as false-matched samples. 

\subsection{Spatial Continuity Index (SDM@K)}
\label{4.2}
\begin{figure*}[!t]
\centering
\includegraphics[width=0.9\textwidth]{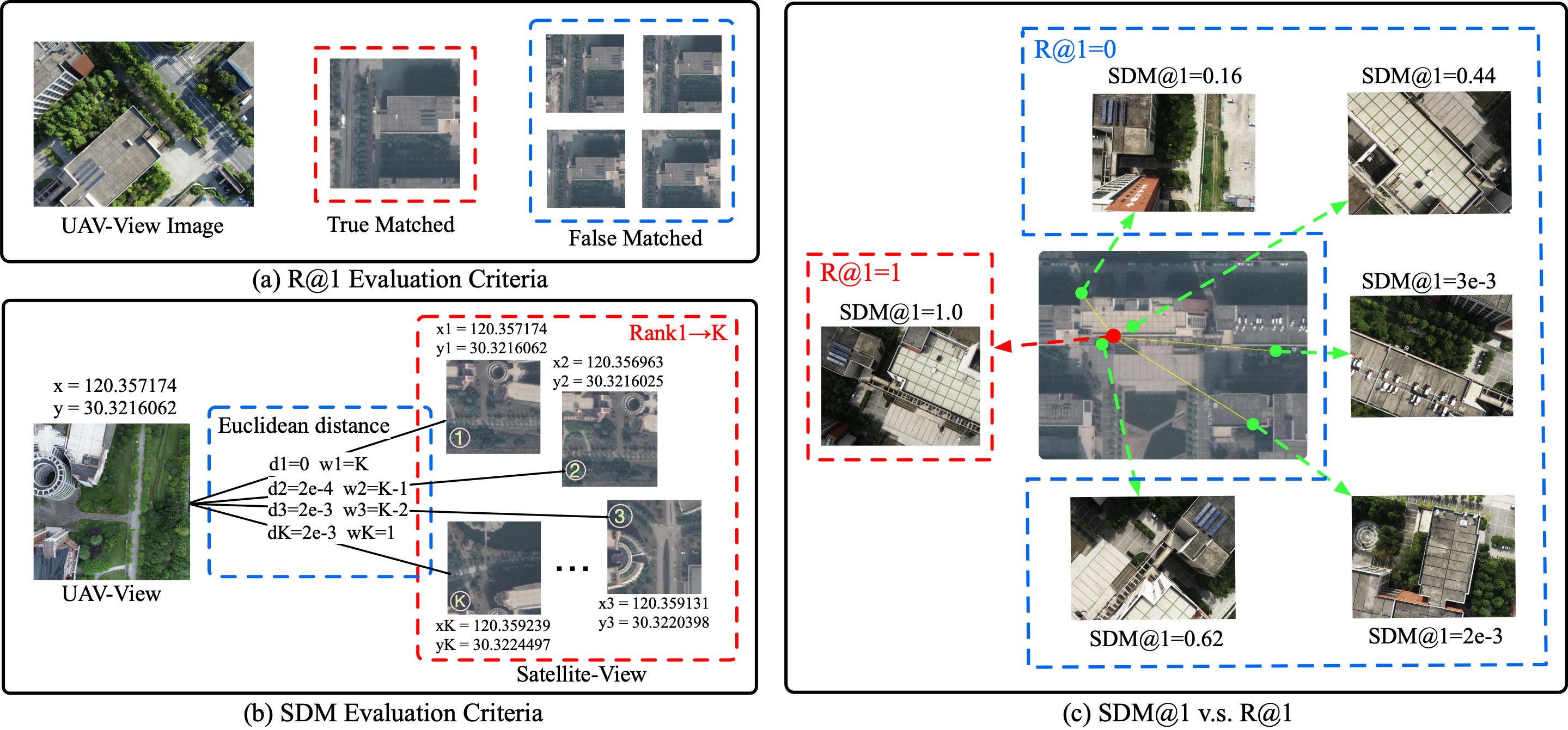}
\caption{A comparison between R@1 and SDM@K metric. In (b), $d1$ to $dK$ represent the spatial euclidean distance between the UAV-view image and the most similar K satellite-view images. The calculation is shown in Equation \ref{eq4}, where $x$ represents the longitude and $y$ represents the latitude. $w1$ to $wK$ are weighting factors and $s$ is an amplification factor. In (c), we compare the evaluation process of R@1 and SDM@1.}
\label{fig_evaluation}
\end{figure*}

During UAV operations, a notable characteristic arises from the spatial continuity and density of satellite imagery, resulting in subtle gaps in feature information between adjacent images.
As depicted in Fig.~\ref{fig_evaluation}~(a), using R@1 as the evaluation criterion would only consider the correct satellite images within the red dashed region, denoted as $I(l_q,l_1)=1$, while categorizing the satellite images within the blue dashed region as incorrect, indicated by $I(l_q,l_1)=0$. 
Despite the presence of minor deviations, the spatial difference is relatively small. 
For UAV self-positioning, the objective is to achieve a closer approximation to the actual location hence minor positioning deviations are inevitable. 
However, large deviations are undesired, and such spatial differences are not effectively captured by the R@1 metric. 

To address the above issue and provide a more accurate measure of localization accuracy, we propose a new evaluation metric, namely Spatial Distance Metric (SDM). 
SDM combines the characteristics of Recall@K while also considering the performance of a model in localization.
Specifically, the SDM value of a single query sample is defined as:
\begin{equation}
	\label{eq3}
	\text{SDM}_k=(K-k+1)/ e^{{s\times{}d}_k}
\end{equation}
where $d_i=\sqrt{{(x_q-x_i)}^2+{(y_q-y_i)}^2}$, $(K-k+1)$ is the weight of the \textit{k-th} sample, as shown in Fig.~\ref{fig_evaluation}~(b). The weight is assigned based on the feature distance.
A large weight is set for the gallery image that is closer to the query feature. 
$K$ denotes the top $K$ samples in the gallery that have the closest distance to the query features. 
$x_q$ and $x_i$ denote the longitude corresponding to the query and gallery images, respectively, while $y_q$ and $y_i$ denote the latitude. 
In short, $d_i$ represents the spatial euclidean distance between two images and $s$ is an amplification factor.
In this paper, $s$ is set to $5\times{10^3}$. 
Last, the SDM values for all $K$ samples are calculated, and the final SDM@K index is obtained through a normalization process:
\begin{equation}
	\label{eq4}
	\text{SDM@K}=\sum_{i=1}^K\frac{(K-i+1)}{e^{{s\times{}d}_i}} / \sum_{i=1}^K(K-i+1).
\end{equation}

It is worth mentioning that the interval of SDM is distributed between 0 and 1.
A larger SDM value indicates a better localization performance of a model. 
The visualization of the calculation process is presented in Fig.~\ref{fig_evaluation}~(b). 
To illustrate this difference more simply, we compare SDM@1 and R@1 in Fig.~\ref{fig_evaluation}~(c).
We can notice that the proposed metric is a continuous evaluation criterion, while R@1 is binary that is not suitable for the proposed UAV self-positioning task. 

Below, the characteristics of the SDM@K indicator are further explained. Overall, SDM is a fusion indicator that combines retrieval and localization tasks. 
In the real positioning process, we allow a little bit of error in positioning, but a large positioning error is not favoured and should be considered a wrong positioning. 
SDM can evaluate this well, which does not define the error like the l1 and l2 distances. 
SDM is not linear but exponentially correlated. The reason for this design is that once the positioning error is large, SDM will give a score close to 0, which is also in line with the retrieval task (belonging to false-matched samples). Specific visualization examples are shown in Fig.~\ref{fig_evaluation}~(c).
SDM divides the specific error situation in the interval with a small positioning error, and it is more sensitive to the error.
Once the error is very large, as shown in the two cases in the bottom right of Fig.~\ref{fig_evaluation}~(c), SDM is almost close to 0 and insensitive. 
In addition, to better evaluate the localization task based on image retrieval, we use both a top-1 sample and as many recalled samples as possible to participate in the evaluation. 
Therefore, SDM uses a hyperparameter of $K$, which represents the top-K images participating in the entire SDM calculation.
The rank of cosine similarity between the recalled image and the query is used as the weight.

\begin{figure*}[!t]%
	\centering
	\includegraphics[width=.95\linewidth]{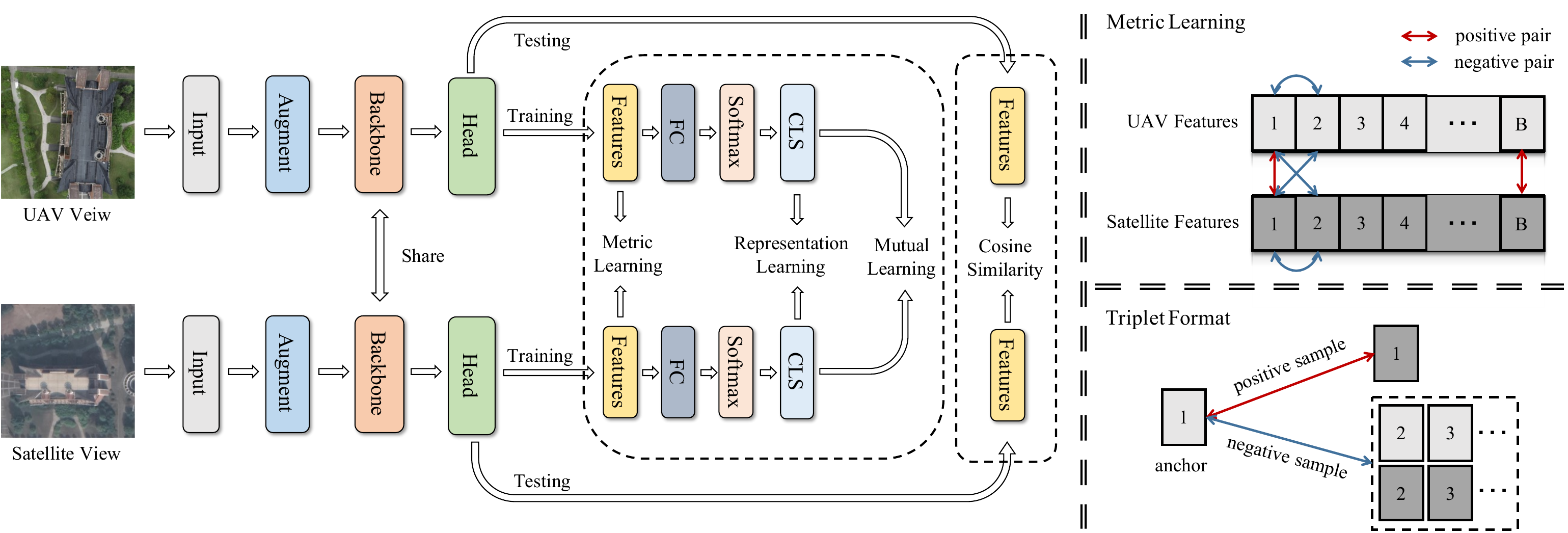}
	\caption{An overview of the proposed framework, including both the training and testing phases.}
	\label{fig_framework}
\end{figure*}

\section{The Proposed Baseline Model}
\label{sec5}
This section will introduce the proposed baseline network for UAV self-positioning, which mainly has four parts: \textit{Augment}, \textit{Backbone}, \textit{Head} and \textit{Loss}. As shown in Fig.~\ref{fig_framework},
first, the model needs to receive the data of the UAV and satellite views as input. 
Then, the input should perform data enhancement through the \textit{Augment} module. 
After the feature extraction step performed in \textit{Backbone}, feature integration is used in the \textit{Head} module and mapped to the specified feature space. 
In the training phase, the feature dimension is mapped to the number of categories by the Fully Connected (FC) layer, and its probability distribution is calculated by the softmax operation. 
Last, a \textit{Loss} function calculates through three types of supervision methods including representation learning, metric learning, and mutual learning. 
During inference, this feature is directly used to calculate the cosine similarity to rank the samples. 
It is worth mentioning that all the weights of the two branches are shared.

\subsection{Data Augmentation}\label{sec5.1}
\begin{figure}[!t]%
	\centering
	\includegraphics[width=0.5\textwidth]{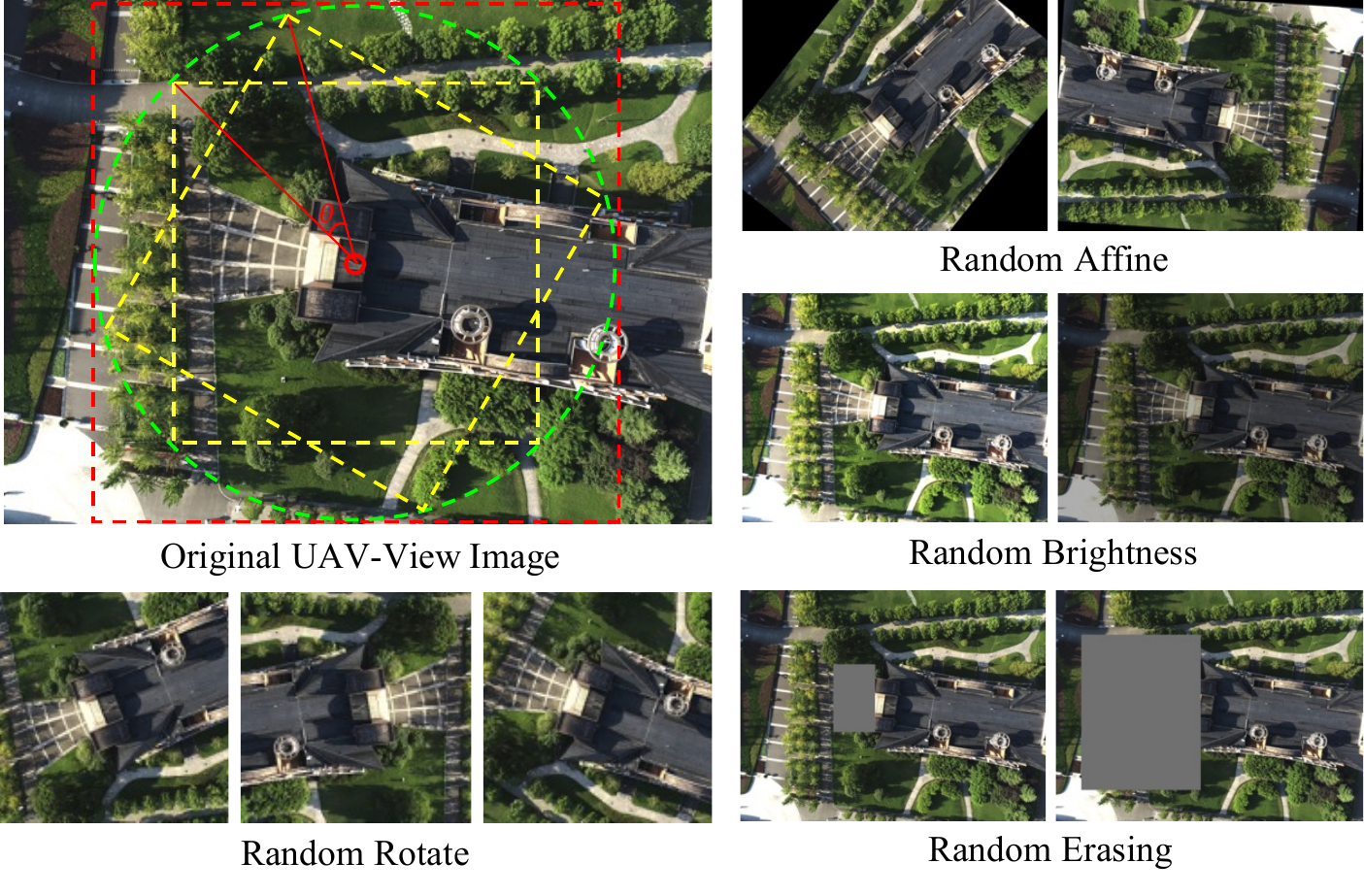}
	\caption{The Visualization of four data augmentation methods. The red dashed line is the largest interior square centered at the image center. The green dashed line is the largest interior circle of the square. The yellow dashed line represents a number of the largest interior squares of the circle. $\theta$ indicates the rotation angle.}
	\label{fig_Aug}
\end{figure}

Given the high cost associated with collecting data using drones in real-world scenes, it becomes particularly meaningful to maximize the diversity of limited training data. Therefore, we adopted four data augmentation methods to expand the UAV and satellite view data.

\subsubsection{UAV Flight Direction}
The unpredictability of the UAV's flight direction presents a challenge during data collection, as it is virtually impossible to capture all possible flight directions. In response to this issue, we propose a novel data augmentation strategy named "random rotate", which is designed to simulate different UAV flight directions. 
Rather than simply rotating the entire original image, random rotate employs a more refined approach. 
In this method, the largest inscribed circle of the original image is identified as the cropping target.
Based on the desired rotation angle $\theta$, a square is extracted from within this circle as depicted in the top of Fig.~\ref{fig_Aug}. 
Random rotate not only expands the data in the UAV's flight direction but also enhances the model's generalization capability by obscuring edge information and allowing the model to focus on more generic features. 
Moreover, by erasing edge information, the model can prioritize central image information. 
The visualization results of three instances of random rotate are shown in the bottom-left corner of Fig.~\ref{fig_Aug}. 

\subsubsection{Satellite View Direction}
Solely applying data augmentation to the UAV direction leads to a rapid convergence of the classification loss for satellite-view images to 100\% during training, indicating a lack of diversity in the satellite data. 
In reality, the orientation of the satellite view is also uncertain. 
Then, we employs the random affine augmentation method for satellite-view data. Unlike random rotate, random affine introduces some empty areas caused by rotation at the image edges, which can introduce some edge noise, thus forcing the model to learn the spatial context information of the satellite imagery.

\subsubsection{Light Intensity}
Images captured by UAVs during different time periods and under varying weather conditions often display variations in brightness. 
To enhance the model's robustness to illumination variations, we introduce the data augmentation method of random brightness. The visualization results are depicted in Fig.~\ref{fig_Aug}. 

\subsubsection{Spatial Differences}
A big challenge in UAV self-positioning is the spatial difference caused by time offset. 
One of the solutions to this challenge is to perform random erasing on images. 
By randomly erasing part of an image, the object changes in space caused by time offset can be indirectly simulated, forcing the model to focus on more general and robust features. 



\subsection{Backbone Network}
\label{sec5.2}
The choice of a proper backbone network plays a crucial role in extracting features from input images. 
The existing backbones mainly include two categories: CNN-based and Transformer-based. In Section \ref{sec6.4}, the effects of different types of backbones on UAV self-positioning are experimentally investigated. Through our experiments, we observed a consistent phenomenon where the performance of the model trained with a CNN-based backbone network is noticeably inferior to that of the model trained with a Transformer-based backbone network. We attribute this discrepancy primarily to the significant challenges at the data level, including domain discrepancies between different perspectives, and variations in spatial information due to perspective and temporal differences. These factors contribute to the model's demand for a strong capability to comprehend global context, which aligns well with the inherent characteristics of Transformer models. To strike a balance between performance and inference speed, we adopt the ViT-S model as the backbone network in our baseline model.

\subsection{Prediction Head}
\label{sec5.3}
The prediction head is responsible for integrating features and compressing them to a specific dimension. In this paper, we divide different prediction heads into two categories: pooling-based and chunking-based. In Section \ref{sec6.5}, we investigate and report the performance of different heads in UAV self-positioning.

\subsubsection{Pooling-based}\label{sec5.3.1}
By default, the output of the ViT model consists of two components: a global class token and other local tokens. Typically, the dimensions of the local tokens are represented as (B, N, C), where B denotes the batch size, N denotes the patch size, and C represents the number of channels. To represent an image with a single feature vector, it is necessary to compress the patch dimension (N) of the local tokens. 
Therefore, various pooling techniques, including MaxPool, AvgPool, AvgMaxPool, and GemPool\cite{gem}, have been evaluated and the results are reported in Section \ref{sec6.5.1}.

\subsubsection{Chunking-based}\label{sec5.3.2}

In recent years, partitioning strategies have been widely used to extract fine-grained features and align regional features. 
One of the representative methods is Local Pattern Network (LPN)\cite{bib18}, which manually divides an image into multiple regions starting from the center and extending towards the boundary of an image. 
Originally, LPN was developed for CNN architectures. 
In this paper, we reconstruct the local tokens obtained from the ViT-S model into a (Height, Width) format, serving as the LPN head input. 
Additionally, we refer to a thermal-based adaptive partitioning method, FSRA\cite{FSRA}, which is designed for ViT structures, eliminating the need for additional operations. 
Detailed analysis and comparison are presented in Section \ref{sec6.5.2}.

\subsection{Loss Functions}\label{sec5.4}
The baseline model encompasses three distinct supervised learning methods: representation learning, metric learning, and mutual learning. These methods need to calculate losses at three different levels, including the class, feature, and distribution levels.

\subsubsection{Representation Learning}\label{sec5.4.1}
Representation learning is a common and efficient method for feature extraction. In essence, representation learning maps data to a specified feature space through a supervision signal. 
The most common one is classification loss, which has been widely used in cross-view geo-localization\cite{JK}, face recognition\cite{face1}, ReID\cite{reid1}, etc. For the baseline model, we use the widely adopted Cross-Entropy (CE) loss\cite{celoss}.

\subsubsection{Metric Learning}\label{sec5.4.2}
During inference, image retrieval tasks often measure the similarity between images based on the distance between their features.
The most two distance measures are the Cosine or Euclidean distances. 
Therefore, directly supervising the model at the feature level during training is essential.
Indeed, metric learning is a such kind of method. 
Generally, metric learning is not used in isolation for supervised learning tasks, but rather in combination with representation learning methods. 
To evaluate the effectiveness of metric learning, we conduct experiments using four types of triplet loss, including the standard \textit{Triplet Loss}\cite{tripletloss}, \textit{Hard-Mining Triplet Loss}, \textit{Same-Domain Triplet Loss}\cite{FSRA}, and \textit{Soft-Weighted Triplet Loss}\cite{SWtripletloss}. 
Based on the experimental results reported in Section \ref{sec6.6.2}, we use the soft-weighted triplet loss as the feature measurement loss function for our baseline model. 
The conventional triplet loss and soft-weighted triplet loss functions are defined as:
\begin{equation}
    \label{eq_tr}
    \text{TriLoss}(a, p, n) = \max(0, D(a, p) \\
    - D(a, n) + m)
\end{equation}
\begin{equation}
    \label{eq_swtr}
    \text{SWTriLoss}(a, p, n) = \log({1+e^{\alpha\times\\ 
        (D(a, p) - D(a, p))}})
\end{equation}
where $a$ is the feature vector of the anchor sample, $p$ is the positive sample feature vector of the anchor sample, and $n$ represents the negative sample feature vector. $m$ is the margin that controls the desired difference in the distance between positive and negative samples. $D(a,b)$ denotes the cosine similarity between sample $a$ and $b$.

\subsubsection{Mutual Learning}\label{sec5.4.3}
Mutual learning has been widely applied to the field of knowledge distillation\cite{Hinton2015DistillingTK}. However, for the UAV self-positioning task, it is expected that the category vector distributions of the UAV and satellite views of the same category tend to be consistent. Therefore, we introduce the distribution-level bilateral learning method, which is expressed as:
\begin{equation}
	\label{eq_8}
	\text{KLLoss} = \text{KLDiv}(O_d||O_s)+\text{KLDiv}(O_s||O_d)
\end{equation}
\begin{equation}
	\label{eq_9}
	\text{KLDiv}(O_p||O_q)=\sum_{i=1}^N{O_p(i)\times\log(\frac{O_p(i)}{O_q(i)})}
\end{equation}
where $O_p$ and $O_q$ respectively represent the probability distribution of the teacher and student category vectors through softmax. Additionally, $O_d$ represents the class vector output of a UAV image, while $O_s$ represents the class vector output of a satellite image.

\section{Experimental Results}\label{sec6}

\subsection{Implementation Details}\label{sec6.1}
The backbone networks used in this section were all pre-trained using the timm framework\cite{rw2019timm}, with the additional classification layer removed. During training, the Stochastic Gradient Descent (SGD) optimizer was employed, with an initial learning rate of 0.003 and batch size of 8. 
The learning rate for the backbone network weights was set to 0.3X of the other weights.
The models were trained for a total of 120 epochs, with the learning rate decreasing to 0.1X at the \textit{70-th} and \textit{110-th} epochs, respectively. 
The input scale for both the UAV-view and satellite-view images was set to 224$\times$224. 
Regarding the network architecture, we use global representation vectors of 512 dimensions through a fully connected layer, followed by a classification layer for category prediction.

\subsection{Non-Dense Features v.s. Dense Features}\label{sec6.2}
Previously, we discussed the significance of dense sampling for UAV self-positioning. In this section, we categorize the dataset into Dense and Non-Dense classes and validate it through experimentation. For the Non-Dense category, we utilize two non-intensive geographic task-related datasets, namely Place\cite{place365} for landmark classification tasks and University-1652 for cross-view geo-localization tasks. For the Dense category, we use the dataset proposed in this paper. The experimental results are shown in Table \ref{tab_dense}, which demonstrate that the use of dense data during training has a significant impact on UAV self-positioning task. Correspondingly, the model trained in dense data can significantly improve the positioning performance. We attribute this improvement to the presence of overlapping areas between adjacent frames in the dense dataset, which compels the model to extract more robust spatial distribution information.

\begin{table}[!t]
	\tiny
	\renewcommand{\arraystretch}{1.0}
	\caption{A comparison of the positioning performance trained with Dense and Non-Dense datasets.}
	\label{tab_dense}
	\resizebox{1.0\hsize}{!}{
		\begin{tabular}{ccccccc}
			\specialrule{0.75pt}{0pt}{1pt}
			{Dataset}& 
			{R@1}&
			{R@5}&
			{SDM@1}&
			{SDM@5}
			\\
			\specialrule{0.5pt}{1pt}{1pt}
			{Place}&
			{1.88\%}&
			{3.22\%}& 
			{4.65\%}&
			{2.80\%}
			\\
			\specialrule{0pt}{0pt}{0pt}
			{University-1652}&
			{25.77\%}&
			{42.63\%}&
			{33.94\%}&
			{21.07\%}
			\\
			\specialrule{0.5pt}{1pt}{1pt}
			{DenseUAV(ours)} &
			\cellcolor{gray!50}80.18\%& 
			\cellcolor{gray!50}93.99\%& 
			\cellcolor{gray!50}84.39\%&
			\cellcolor{gray!50}78.02\%
			\\
			\specialrule{0.75pt}{1pt}{0pt}
		\end{tabular}
	}
\end{table}

\subsection{Evaluation on Data Augmentation}
\label{sec6.3}
It is crucial to create diverse scenarios that better align with real-world conditions. This section focuses on the impact of data augmentation techniques on UAV self-positioning task.

\begin{table*}[h]
	\tiny
	\centering
	\renewcommand\arraystretch{1.05}
	\caption{The experiment results obtained by different data augmentation configurations.}
	\label{table_Aug}
	\resizebox{0.9\hsize}{!}{
		\begin{tabular}{cc|cc|cc|cc|ccccc}
			\specialrule{0.75pt}{0pt}{1pt}
			\multicolumn{2}{c|}{Random Rotate} & \multicolumn{2}{c|}{Random Affine} &
			\multicolumn{2}{c|}{Random Brightness} & \multicolumn{2}{c|}{Random Erasing} & 
			\multirow{2}{*}{R@1} & \multirow{2}{*}{SDM@1} & \multirow{2}{*}{SDM@3} & \multirow{2}{*}{SDM@5} & \multirow{2}{*}{SDM@10} \\
			{Drone} & {Satelite} & {Drone} & {Satelite} & {Drone} & {Satelite} & {Drone} & {Satelite} \\
			\specialrule{0.5pt}{1pt}{1pt}
			&  &  &  &  &  &  &  & 67.57\% & 74.03\% & 71.73\% &67.67\% & 58.19\% \\
			\specialrule{0.5pt}{1pt}{1pt}
			\checkmark &  &  &  &  &  &  &  & \cellcolor{gray!50}76.36\% & \cellcolor{gray!50}80.60\% & \cellcolor{gray!50}78.18\% &\cellcolor{gray!50}73.71\% & 62.39\% \\
			& \checkmark &  &  &  &  &  &  & 73.57\% & 78.66\% & 76.85\% &73.05\% & \cellcolor{gray!50}62.88\% \\
			\checkmark& \checkmark &  &  &  &  &  &  & 72.46\% & 76.70\% & 74.21\% &69.74\% & 58.86\% \\
			\specialrule{0.5pt}{1pt}{1pt}
			\checkmark&  & \checkmark & \checkmark &  &  &  &  & 75.68\% & 80.39\% & 78.47\% &74.08\% & 62.80\% \\
			\checkmark&  &  & \checkmark &  &  &  &  & \cellcolor{gray!50}77.65\% & \cellcolor{gray!50}81.88\% & \cellcolor{gray!50}80.00\% &\cellcolor{gray!50}75.60\% & \cellcolor{gray!50}64.39\% \\
			\specialrule{0.5pt}{1pt}{1pt}
			\checkmark&  &  & \checkmark & \checkmark &  &  &  & 76.66\% & 81.18\% & 79.47\% &75.28\% & 64.25\% \\
			\checkmark&  &  & \checkmark &  & \checkmark &  &  & 76.45\% & 81.15\% & 79.21\% &75.07\% & 64.25\% \\
			\checkmark&  &  & \checkmark & \checkmark & \checkmark &  &  & \cellcolor{gray!50}77.43\% & \cellcolor{gray!50}81.70\% & \cellcolor{gray!50}79.84\% &\cellcolor{gray!50}75.61\% & \cellcolor{gray!50}64.33\% \\
			\specialrule{0.5pt}{1pt}{1pt}
			\checkmark&  &  & \checkmark &  &  & \checkmark &  & 78.42\% & 82.84\% & 81.02\% &76.57\% & 65.29\% \\
			\checkmark&  &  & \checkmark &  &  &  & \checkmark & \cellcolor{gray!50}80.18\% & \cellcolor{gray!50}84.39\% & 82.51\% &78.02\% & 66.46\% \\
			\checkmark&  &  & \checkmark &  &  & \checkmark & \checkmark & 79.84\% & 84.16\% & \cellcolor{gray!50}82.92\% &\cellcolor{gray!50}78.91\% & \cellcolor{gray!50}67.32\% \\
			
			\specialrule{0.75pt}{1pt}{0pt}
		\end{tabular}
	}
\end{table*}

\subsubsection{UAV Flight Direction}\label{sec6.3.1}

To validate the efficacy of our proposed random rotate, we conducted experiments as presented in the \textit{Random Rotate} column of Table \ref{table_Aug}. 
The first row in the table represents the use of only random horizontal flip and resize operations at the data level.
We can notice that incorporating random rotate data augmentation for the UAV view led to significant improvements across all the metrics.
For example, the R@1 metric is increased by nearly 9\% and the SDM@1 metric is improved 6\%. This also fully reflects the importance of UAV flight direction data expansion, and also verifies the effectiveness of the proposed random rotate method for flight direction data expansion.

\subsubsection{Satellite View Direction}\label{sec6.3.2}
The impact of random affine has also been evaluated in the \textit{Random Affine} column of Table \ref{table_Aug}. 
The application of random affine for satellite-view images has shown improved performance across various metrics. Specifically, the R@1 metric has increased by 1.3\%, and the SDM@1 metric has increased by 1.2\%. However, when applied to UAV view images, there is a slight performance degradation observed after implementing random affine. This can be attributed to the fact that random affine is not well-suited for accurately capturing the real scene of drone rotation.

\subsubsection{Light Intensity}\label{sec6.3.3}
The impact of light intensity has also been evaluated in the \textit{Random Brightness} column of Table \ref{table_Aug}. 
The results demonstrate that random brightness has minimal impact on the results. 
This can be attributed to the random time-based data collection, which already encompasses data samples with varying brightness levels. It can also be concluded that lighting is not the main challenge and factor affecting this task.

\subsubsection{Spatial Difference}\label{sec6.3.4}
The impact of spatial difference has also been evaluated in the \textit{Random Erasing} column of Table \ref{table_Aug}. 
On the whole, random erasing generally improves positioning accuracy. In particular, the effect on satellite view is more significant, increasing R@1 by about 2.5\% and SDM@1 by about 2.5\%.
This is mainly due to the fact that random erasing can create a scene where information is lost in space, which can help the model effectively overcome the challenge of spatial information changes caused by time offsets.

\subsection{Evaluation on Backbone Network}\label{sec6.4}
\begin{figure}[!t]%
	\centering
	\includegraphics[width=1\linewidth]{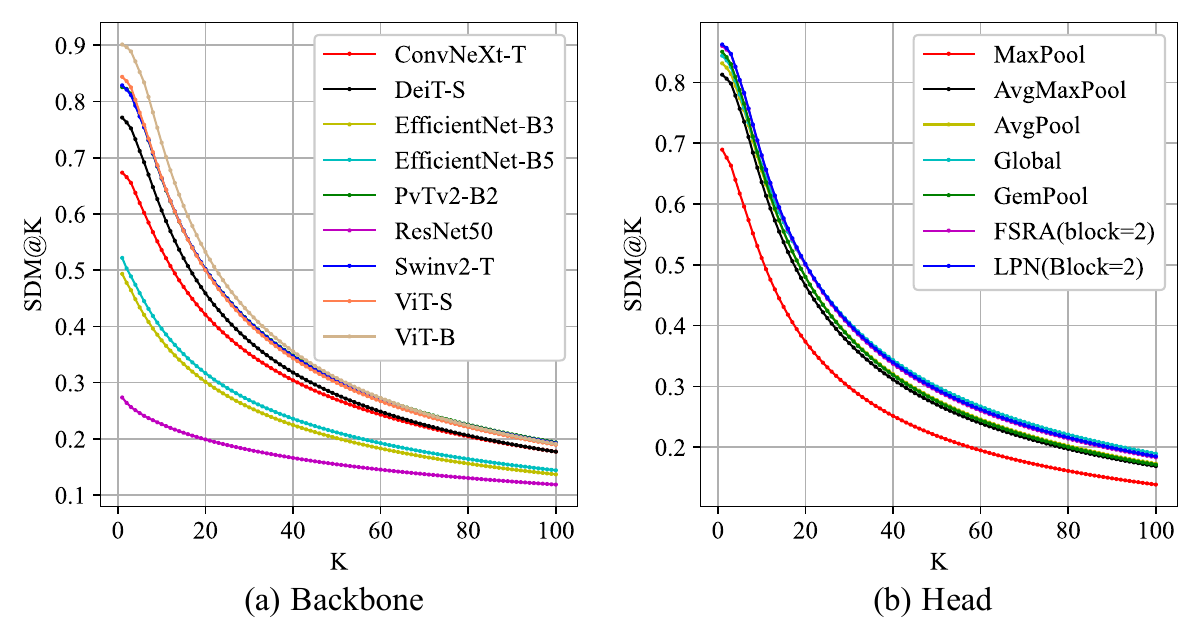}
	\caption{SDM curve diagram. (a) is the SDM curve diagram of different backbones, (b) is the SDM curve diagram of different heads}
	\label{fig_SDM_curve_backbone_head}
\end{figure}

To explore the impact of different backbone networks on the UAV self-positioning task, we adopt some popular backbone networks for experiments, mainly containing two categories: CNN-based and Transformer-based. Among them, the ResNet50, EfficientNet-B3, EfficientNet-B5, and ConvNeXt-T models are used as CNN backbones. 
DeiT-S, PvTv2-B2, Swinv2-T, ViT-S, and ViT-B are used as Transformer backones. 
Note that the Swinv2-T model uses an input size of 256$\times$256, and all the other networks use 224$\times$224 inputs. 
All the above models are trained on the DenseUAV dataset. 
The weights of all models are pre-trained with Timm \cite{rw2019timm}. 
The experimental results are shown in Table~\ref{table_Backbone}, including the number of parameters (Params), calculation amount (Macs), inference time (InferTime, for 2000 samples), and retrieval accuracy (R@1 and R@5). 
In addition, the curve of SDM@K is plotted in Fig.~\ref{fig_SDM_curve_backbone_head}~(a). 

Based on the experimental results, the following conclusions are drawn.
First, the Transformer-based models perform significantly better than the CNN-based models for UAV self-positioning. 
Second, ViT, the most primitive Transformer model, is more competitive. 
Last, in terms of both accuracy and inference speed, the ViT-S model reaches a good balance, which achieves 80.18\%/84.39\% in R@1/SDM@1, with only 18.9s/2000images in inference time.

\begin{table}[!t]
	\centering
	\tiny
	\renewcommand\arraystretch{1.05}
	\caption{A comparison of different backbone networks, including the number of parameters, computation, inference speed, and R@K accuracy.}
	\label{table_Backbone}
	\resizebox{1.0\hsize}{!}{
		\begin{tabular}{c|ccc|cc}
			\specialrule{0.75pt}{0pt}{1pt}
			{Backbone} & {Params} & {Macs} & {InferTime} & {R@1} & {R@5}\\
			\specialrule{0.5pt}{1pt}{1pt}
			ResNet50\cite{resnet} & 27.8M & 8.2G & 20.4s & 16.52\% & 39.30\%  \\
			\specialrule{0pt}{0pt}{0pt}
			EfficientNet-B3\cite{efficientnet} & 14.1M & 1.9G & 46.8s & 42.81\% & 64.52\%  \\
			\specialrule{0pt}{0pt}{0pt}
			EfficientNet-B5\cite{efficientnet} & 32.3M & 4.7G & 67.7s & 44.96\% & 67.78\%  \\
			\specialrule{0pt}{0pt}{0pt}
			ConvNeXt-T\cite{convnext} & 30.1M & 8.9G & 16.9s & \cellcolor{gray!50}60.23\% & \cellcolor{gray!50}81.94\%  \\
			\specialrule{0.5pt}{1pt}{1pt}
			DeiT-S\cite{deit} & 23.7M & 8.5G & 19.2s & 71.77\% & 89.70\%  \\
			\specialrule{0pt}{0pt}{0pt}
			PvTv2-B2\cite{pvt} & 26.8M & 7.8G & 40.9s & 77.99\% & 92.79\% \\
			\specialrule{0pt}{0pt}{0pt}
			Swinv2-T\cite{swinv2} & 29.9M & 8.8G & 38.5s & 77.99\% & 92.49\% \\
			\specialrule{0pt}{0pt}{0pt}
			ViT-S\cite{bib20} & 23.3M & 8.5G & 18.9s & 80.18\% & 93.99\% \\
			\specialrule{0pt}{0pt}{0pt}
			ViT-B\cite{bib20} & 87.9M & 33.7G & 20.2s & \cellcolor{gray!50}87.82\% & \cellcolor{gray!50}97.17\% \\
			\specialrule{0.75pt}{1pt}{0pt}
		\end{tabular}
	}
\end{table}

\begin{figure}[!t]%
	\centering
	\includegraphics[width=1\linewidth]{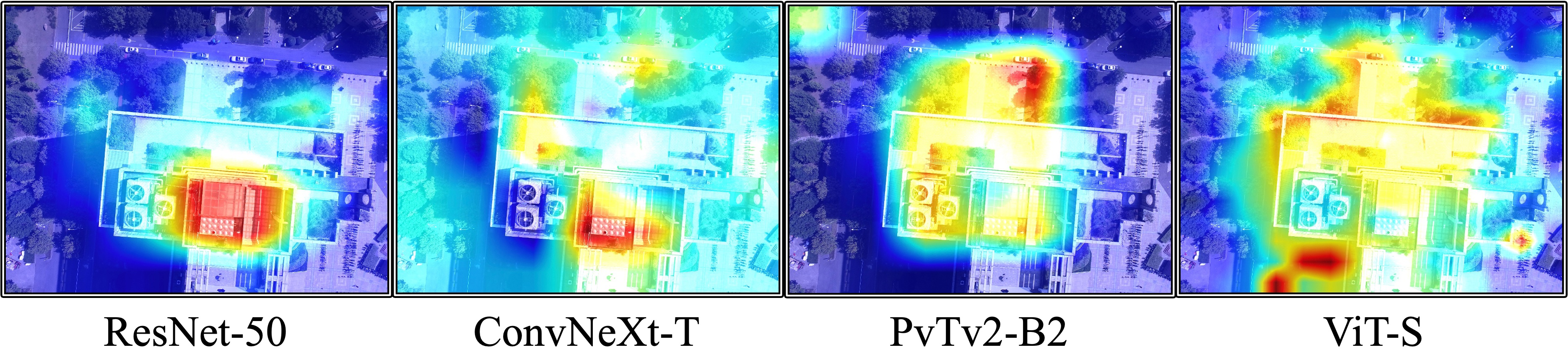}
	\caption{The thermal distribution perspective visualizes the spatial response of 4 different types of backbone networks.}
	\label{fig_heatmap}
\end{figure}

It can be observed that there is a significant gap in positioning accuracy between CNN-based and Transformer-based methods.
To further explore the advantages of Transformers, we visualize the heatmap results of four different types of backbone networks in Fig.~\ref{fig_heatmap}.
Due to its limited receptive field, ResNet-50 focuses mainly on salient feature regions, which is disadvantageous for UAV self-positioning since adjacent frames often have overlapping areas. 
By solely focusing on salient feature regions, it becomes difficult to achieve accurate self-positioning results. 
Unlike ResNet, ConvNeXt, with the addition of large receptive field convolutions in its network structure, no longer restricts the attention to salient regions. 
However, from the heatmap, it can be observed that the attention is too dispersed. 
However, for the Transformer models, either the simplest ViT model or the hierarchical PvT model, their attention is related to the spatial distribution of images. 
The PvTv2-B2 model even prioritizes the central region of an image, which aligns well with the intention of UAV self-positioning (the UAV's position is typically at the center of an image).
However, this approach ignores global information and the importance of spatial context.
In contrast, the ViT-S model preserves the prioritization of the central region in terms of spatial heatmap distribution, while also focusing on spatial contextual semantic information.

\subsection{Evaluation on Prediction Head}
\label{sec6.5}
The role of a prediction head is to integrate and map the features extracted from the backbone. This paper divides the heads into two groups.
The first one is based on pooling methods, and the other one is based on some existing chunking methods. 
Since the chunking-based method has a limited convergence speed under the condition of the default learning rate, to present the results more fairly, the learning rate of the chunking-based method is set to 0.01 (the default is 0.003). 
In addition, all the experiments in this section use the ViT-S model as the backbone network.
The results are reported in Table \ref{table_Head} and Fig.~\ref{fig_SDM_curve_backbone_head}~(b), including the parameters, computation, inference time, and accuracy of different heads.

\begin{table}[!t]
	\centering
\tiny
	\renewcommand\arraystretch{1.05}
	\caption{A comparison of different heads in parameters, computation, inference speed, and R@K accuracy.}
	\label{table_Head}
	\resizebox{1.0\hsize}{!}{
		\begin{tabular}{c|ccc|cc}
			\specialrule{0.75pt}{0pt}{1pt}
			{Head} & {Params} & {Macs} & {InferTime} & {R@1} & {R@5}\\
			\specialrule{0.5pt}{1pt}{1pt}
			MaxPool & 23.3M & 8.4G & 18.9s & 65.21\% & 82.45\%  \\
			AvgPool & 23.3M & 8.4G & 19.0s & 79.62\% & 92.96\% \\
			AvgMaxPool & 23.3M & 8.4G & 19.3s & 77.95\% & 93.35\%    \\
			Global (baseline) & 23.3M & 8.4G & 18.9s & 80.18\% & 93.99\% \\
			GemPool\cite{gem} & 23.3M & 8.4G & 19.8s & 82.45\% & 94.21\%  \\
			\specialrule{0.5pt}{1pt}{1pt}
			FSRA(Block=2)\cite{FSRA} & 26.0M & 8.5G & 21.1s & 82.58\% & \cellcolor{gray!50}94.94\% \\
			LPN(Block=2)\cite{bib18} & 26.0M & 8.5G & 21.2s & \cellcolor{gray!50}83.05\% & 94.89\% \\
			
			\specialrule{0.75pt}{1pt}{0pt}
		\end{tabular}
	}
\end{table}

\subsubsection{Pooling-based Heads}
\label{sec6.5.1}
Table \ref{table_Head} shows the results of different heads. Among them, MaxPool indicates that the 1D max pooling operation is performed in the patches dimension of local tokens. 
Similarly, AvgPool represents a 1D average pooling operation. 
AvgMaxPool is the sum of the results of MaxPool and AvgPool. 
Global means just using the global class token. 
GemPool means to adaptively learn weights for local tokens through learnable parameters. 
Global is the baseline model used in this paper, which has a certain improvement as compared with AvgPool and MaxPool. 
Due to its adaptive learning method, GemPool has improved R@1 by 2.3\% and SDM@1 by 0.6\% as compared with the baseline method.

\subsubsection{Chunking-based Heads}\label{sec6.5.2}
The main idea of the chunking-based method is to achieve feature alignment between images from different perspectives by blocks. To maintain alignment, the number of blocks is uniformly set to 2. As shown in Table \ref{table_Head}, compared with the baseline, both LPN and FSRA have improved the final performance significantly. 
Among them, LPN has improved R@1 by 2.9\%, and SDM@1 by 1.8\%. 
However, FSRA and LPN increase the inference time slightly.

\subsection{Evaluation on Loss Function}
\label{sec6.6}
\begin{table*}[h]
	\centering
	\renewcommand\arraystretch{1.05}
	\caption{A comparison of different loss functions in localization performance.}
	\label{table_Loss}
	\resizebox{1.0\hsize}{!}{
		\begin{tabular}{cc|cccc|c|ccccc}
			\specialrule{1.5pt}{0pt}{2pt}
			\multicolumn{2}{c|}{Representation Learning} & 
			\multicolumn{4}{c|}{Metric Learning} &
			\multicolumn{1}{c|}{Mutual Learning} &
			{\makecell[c]{\multirow{2}{*}[-2ex]{R@1}}} & 
			{\makecell[c]{\multirow{2}{*}[-2ex]{SDM@1}}} & 
			{\makecell[c]{\multirow{2}{*}[-2ex]{SDM@3}}} & 
			{\makecell[c]{\multirow{2}{*}[-2ex]{SDM@5}}} & 
			{\makecell[c]{\multirow{2}{*}[-2ex]{SDM@10}}}
			\\
			\cmidrule[0.5pt](rl){1-7}
			{\makecell[c]{CE Loss}} & 
			{\makecell[c]{Focal Loss}} & 
			{\makecell[c]{Triplet Loss}} & 
			{\makecell[c]{Hard-Mining\\Triplet Loss}} & 
			{\makecell[c]{Same-Domain\\Triplet Loss}} & 
			{\makecell[c]{Soft-Weighted\\Triplet Loss}} & 
			{KL Loss} 
			\\
			\specialrule{0.5pt}{2pt}{2pt}
			\checkmark&  &  &  &  &  &  & \cellcolor{gray!50}76.28\% & \cellcolor{gray!50}81.23\% & \cellcolor{gray!50}79.25\% & \cellcolor{gray!50}74.39\% & \cellcolor{gray!50}63.00\% \\
			& \checkmark &  &  &  &  &  & 73.19\% & 79.14\% & 77.04\% & 72.46\% & 61.98\% \\
			\specialrule{0.5pt}{2pt}{2pt}
			\checkmark&  & \checkmark &  &  &  &  & 79.19\% & 83.63\% & 81.20\% & 76.90\% & 65.26\% \\
			\checkmark&  &  & \checkmark &  &  &  & 80.18\% & 84.39\% & 82.51\% & 78.02\% & 66.46\% \\
			\checkmark&  &  &  & \checkmark &  &  & 81.17\% & 85.09\% & 82.93\% & 78.51\% & 67.00\% \\
			\checkmark&  &  &  &  & \checkmark &  & \cellcolor{gray!50}81.17\% & \cellcolor{gray!50}85.22\% & \cellcolor{gray!50}83.10\% & \cellcolor{gray!50}78.89\% & \cellcolor{gray!50}67.26\% \\
			\specialrule{0.5pt}{2pt}{2pt}
			\checkmark&  &  &  &  & \checkmark & \checkmark & \cellcolor{gray!50}83.01\% & \cellcolor{gray!50}86.50\% & \cellcolor{gray!50}84.50\% & \cellcolor{gray!50}80.44\% & \cellcolor{gray!50}68.49\% \\
			\specialrule{1.5pt}{2pt}{0pt}
		\end{tabular}
	}
\end{table*}
Our baseline includes three supervised methods: \textit{Representation Learning}, \textit{Metric Learning}, and \textit{Mutual Learning}.

\subsubsection{Representation Learning}\label{sec6.6.1}

In the experiments, we evaluate two commonly used classification losses: \textit{Cross-Entropy (CE) Loss} and \textit{Focal Loss}\cite{focalloss}. The results are shown in Table \ref{table_Loss}. Although focal loss provides advantages in terms of sample and weight balancing, we find that the CE loss is more suitable for the task of UAV self-positioning.

\subsubsection{Metric Learning}\label{sec6.6.2}
Metric learning serves the enhancement of the discriminative capability of the model and mitigates modality discrepancies. 
To verify the effectiveness of metric learning, we perform experiments using four variations of triplet loss: \textit{Triplet Loss}, \textit{Hard-Mining Triplet Loss}, \textit{Same-Domain Triplet Loss}, and \textit{Soft-Weighted Triplet Loss}. 
As shown in Table \ref{table_Loss}, compared to using only representation learning, the inclusion of triplet loss leads to a noticeable improvement of 2.9\% in R@1 and an increase of 2.4\% in SDM@1. 
Furthermore, the soft-weighted triplet loss achieves an additional performance boosting of 2\% in R@1 and 1.6\% in SDM@1, as compared with the standard triplet loss.

\subsubsection{Mutual Learning}\label{sec6.6.3}
Representation learning and metric learning are widely used techniques in feature learning. 
In contrast, Kullback-Leibler (KL) divergence is primarily applied in knowledge distillation scenarios, particularly in teacher-student learning settings. 
In this paper, we adopt a bidirectional mutual learning approach, where two branches exchange and enhance their knowledge and representations. 
The mutually supervised learning approach aids in ensuring that the UAV and satellite branches progress simultaneously and mutually influence each other. 
In our experiment, we introduce KLLoss based on the CE loss and soft-weighted triplet loss.
As shown in Table \ref{table_Loss}, the use of KLLoss leads to further performance boosting, \textit{i.e.,} 1.8\% in R@1 and 1.2\% in SDM@1.

\subsection{The Impact of Data Source}
\label{sec6.7}
There are many factors that affect the performance of the UAV self-positioning task. Therefore, this section will combine experiments and analyze the impact of data sources, mainly including flight height, the scale, and the time node of the satellite image.

\begin{table*}[h]
	\centering
	\tiny
	\renewcommand\arraystretch{1.0}
	\caption{The impact of some parameters of the data source on the positioning ability of UAV.}
	\label{table_datatype}
	\resizebox{1.0\hsize}{!}{
		\begin{tabular}{ccc|ccc|cc|cccccc}
			\specialrule{0.75pt}{0pt}{2pt}
			\multicolumn{3}{c|}{Drone-View} & 
			\multicolumn{5}{c|}{Satellite-View} &
			{\makecell[c]{\multirow{3}{*}[-3ex]{R@1}}} & 
			{\makecell[c]{\multirow{3}{*}[-3ex]{R@5}}} & 
			{\makecell[c]{\multirow{3}{*}[-3ex]{SDM@1}}} & 
			{\makecell[c]{\multirow{3}{*}[-3ex]{SDM@3}}} & 
			{\makecell[c]{\multirow{3}{*}[-3ex]{SDM@5}}} & 
			{\makecell[c]{\multirow{3}{*}[-3ex]{SDM@10}}}
			\\
			\cmidrule[0.5pt](rl){1-8}
			\multicolumn{3}{c|}{Height} & 
			\multicolumn{3}{c|}{Scale} &
			\multicolumn{2}{c|}{Time}
			\\
			\cmidrule[0.5pt](rl){1-8}
			{\makecell[c]{80m}} & 
			{\makecell[c]{90m}} & 
			{\makecell[c]{100m}} & 
			{\makecell[c]{Small}} & 
			{\makecell[c]{Middle}} & 
			{\makecell[c]{Big}} & 
			{\makecell[c]{2020}} & 
			{\makecell[c]{2022}}
			\\
			
			\specialrule{0.5pt}{2pt}{2pt}
			
			\checkmark&&& \checkmark&\checkmark&\checkmark&\checkmark&\checkmark&79.70\%&94.33\%&84.07\%&82.83\%&\cellcolor{gray!50}78.60\%&\cellcolor{gray!50}67.19\%
			\\
			&\checkmark&& \checkmark&\checkmark&\checkmark&\checkmark&\checkmark&80.33\%&93.15\%&84.12\%&81.85\%&77.26\%&65.61\%
			\\
			&&\checkmark& \checkmark&\checkmark&\checkmark&\checkmark&\checkmark&\cellcolor{gray!50}80.82\%&\cellcolor{gray!50}94.63\%&\cellcolor{gray!50}85.25\%&\cellcolor{gray!50}83.10\%&78.49\%&66.77\%
			\\
			\specialrule{0.5pt}{2pt}{2pt}
			\checkmark&\checkmark&\checkmark& \checkmark&&&\checkmark&\checkmark&76.36\%&95.50\%&81.42\%&66.90\%&55.79\%&40.26\%
			\\
			\checkmark&\checkmark&\checkmark& &\checkmark&&\checkmark&\checkmark&79.28\%&96.57\%&83.41\%&68.36\%&56.81\%&40.90\%
			\\
			\checkmark&\checkmark&\checkmark& &&\checkmark&\checkmark&\checkmark&76.49\%&96.05\%&81.55\%&66.33\%&55.12\%&39.53\%
			\\
			\checkmark&\checkmark&\checkmark& \checkmark&\checkmark&&\checkmark&\checkmark&\cellcolor{gray!50}80.57\%&\cellcolor{gray!50}96.98\%&\cellcolor{gray!50}84.52\%&79.24\%&72.32\%&57.27\%
			\\
			\checkmark&\checkmark&\checkmark& &\checkmark&\checkmark&\checkmark&\checkmark&79.15\%&94.81\%&83.61\%&78.89\%&71.97\%&56.94\%
			\\
			\checkmark&\checkmark&\checkmark& \checkmark&\checkmark&\checkmark&\checkmark&\checkmark&80.18\%&93.99\%&84.39\%&\cellcolor{gray!50}82.51\%&\cellcolor{gray!50}78.02\%&\cellcolor{gray!50}66.46\%
			\\
			\specialrule{0.5pt}{2pt}{2pt}
			\checkmark&\checkmark&\checkmark& \checkmark&\checkmark&\checkmark&\checkmark&&71.99\%&90.69\%&76.66\%&74.85\%&66.01\%&49.71\%
			\\
			\checkmark&\checkmark&\checkmark& \checkmark&\checkmark&\checkmark&&\checkmark&70.48\%&87.00\%&75.79\%&73.38\%&64.59\%&49.10\%
			\\
			\checkmark&\checkmark&\checkmark& \checkmark&\checkmark&\checkmark&\checkmark&\checkmark&\cellcolor{gray!50}80.18\%&\cellcolor{gray!50}93.99\%&\cellcolor{gray!50}84.39\%&\cellcolor{gray!50}82.51\%&\cellcolor{gray!50}78.02\%&\cellcolor{gray!50}66.46\%
			\\
			\specialrule{0.75pt}{2pt}{0pt}
		\end{tabular}
	}
\end{table*}

\subsubsection{Impact of Flight Altitude}

The altitude at which a UAV operates directly impacts the visible range of its captured UAV-view images. Furthermore, increasing the flight altitude can expands the field of view, but it also leads to a larger spatial distance represented by each pixel after image resizing, resulting in a loss of fine-grained information. To investigate the impact of various flight altitudes, we conduct experiments by using the captured UAV images at heights of 80m, 90m, and 100m. 

From the experimental results, we can see that with the increase of the flying height of the drone, the indicators \textit{i.e.,} R@1 and SDM@1 show an increasing trend, mainly because the flying height of the drone can make the image contain a wider field of view, that is, contain more abundant spatial information. However, due to the limited flying height of drones, we were unable to collect data at higher altitudes. The impact of higher UAV flight altitudes on positioning performance still needs to be further explored.

\subsubsection{Impact of the Scale}
Intuitively, the scale of the satellite image determines the amount of information contained in the satellite-view images. If the scale is too small, it will increase the difficulty of matching, and if the scale is too large, it will increase the probability of wrong matching. 
Multi-scale satellite imagery can help the model to improve its robustness to scale changes in the UAV input. 
To explore the influence of satellite-view scale on UAV self-positioning, this section conducts comparative experiments on satellite images of different scales, and the results are shown in the Scale column of Table \ref{table_datatype}. 
Among them, the definition of Small, Middle, and Big is as follows. If Middle is used as the base scale (B, B), the scale of Small is (0.75B, 0.75B), and the scale of Big is (1.25B, 1.25B). 
First, it can be found that in the single scale set, the middle scale achieves the best performance, and the performance of the small and big are similar, which is lower than middle-scale by nearly 3 points in R@1. This is mainly because middle-scale images contain a moderate amount of spatial scale, which is more suitable for query images.
Then, we can find that multi-scale has stronger performance than single-scale universally. Specifically, the simultaneous use of the total three scales, as compared to the use middle scale merely, results in a 0.9\% improvement in R@1 and 1.0\% in SDM@1. This also makes us think about a problem. In real landing application scenarios, the flying height of the UAV is unknown, so the satellite image library should contain images of as many scales as possible to make up for the uncertainty of the UAV’s flying height.

\subsubsection{Impact of the Time}
The spatial difference arising from time offset poses a significant challenge in UAV self-positioning. To enhance the model's robustness to this type of difference, this paper constructed the dataset using satellite-view images from two different years (2020 and 2022), while the UAV-view images were acquired in 2021. As shown in the Time column of Table \ref{table_datatype}.
The experimental results demonstrate that incorporating multi-time spans yields better results compared to a single-time node. When compared to a single time node, R@1 improves by nearly 9\%, and SDM@1 increases by approximately 7 points. These results highlight the importance of considering multi-time spans in addressing the spatial difference caused by time offset. By including satellite images from different years, the model becomes more robust to temporal variations and exhibits improved performance in UAV self-positioning tasks.

\subsection{Visualization}
\label{sec6.8}
\begin{figure}[!t]%
	\centering
	\includegraphics[width=0.5\textwidth]{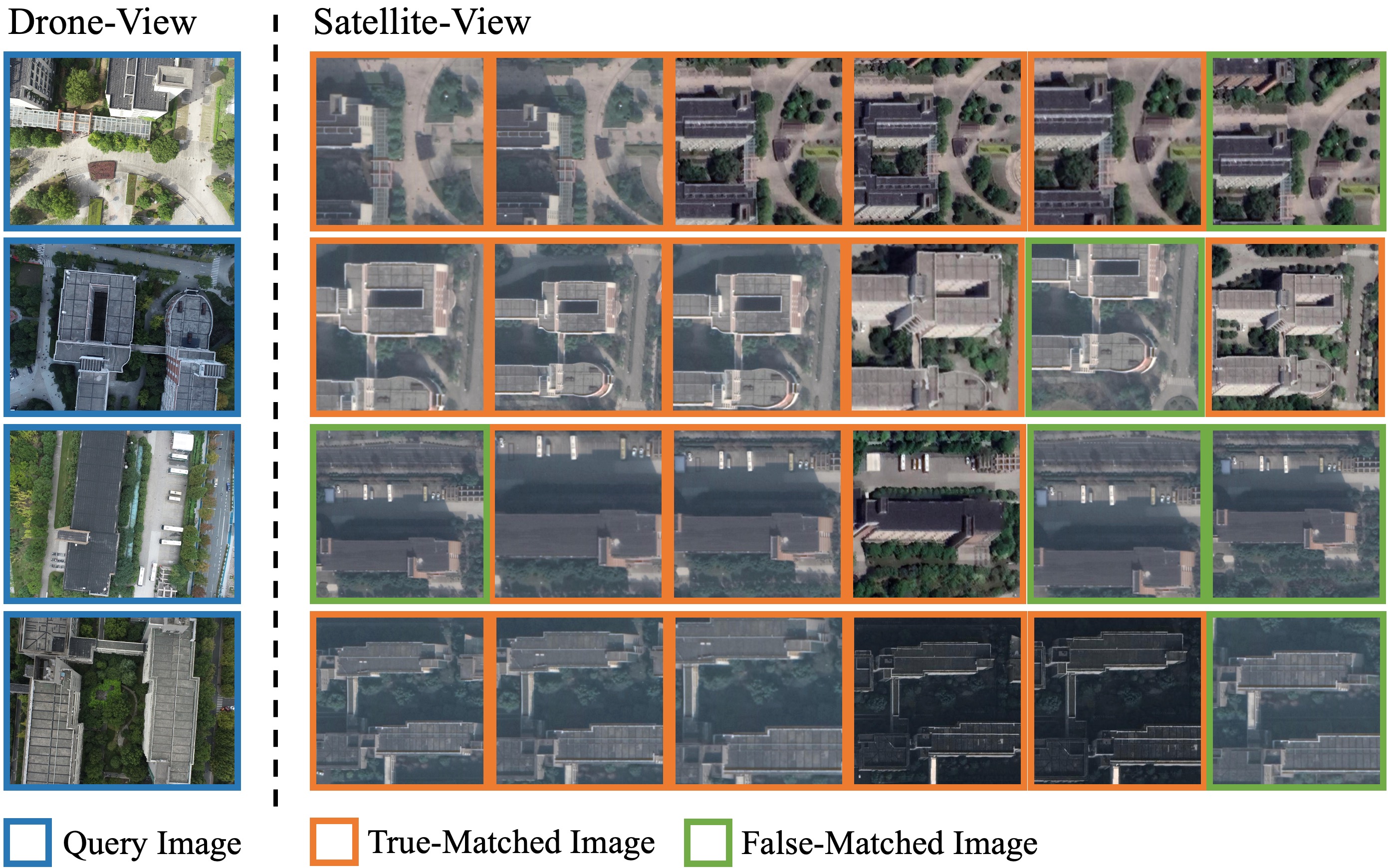}
	\caption{Some results obtained by the baseline model.}
	\label{fig_sample_visualization}
\end{figure}

This subsection visualizes the positioning results of the baseline model from the sample level. 
Fig.~\ref{fig_sample_visualization} illustrates the retrieval results of 4 sample groups within the test set. The UAV-view images (query) are depicted on the left side of the dotted line, while the 6 images closest to the query image in the satellite view are displayed on the right side. 
The true-matched images are denoted by orange boxes, while the false-matched images are represented by green boxes. 
Notably, satellite images encompass a span of time, leading to substantial domain variations, whereas UAV images exhibit discrepancies in brightness and shooting height due to temporal inconsistencies. 
As depicted in the figure, there is a noticeable divergence in both domain and spatial information between UAV and satellite images, posing a significant challenge in this task. Although this paper adopts the method of metric learning and mutual learning to alleviate the matching difficulties caused by domain differences. Some queries still cannot match the corresponding satellite images well. Perhaps some perspective transformation and GAN methods can also be used to alleviate this domain difference.

\section{Conclusion}
\label{sec7}
This study investigated the vision-based self-positioning task of UAVs in low-altitude urban environments. Firstly, we introduced DenseUAV, a novel UAV-based geo-localization dataset, which incorporates real-world scene acquisition and dense sampling. 
This dataset aims to establish a strong pipeline between UAV high-precision self-positioning and real-world applications.
Furthermore, we proposed a new evaluation metric, SDM@K, which offers enhanced effectiveness in evaluating positioning accuracy within real-world scenarios.
Additionally, we developed a robust baseline model and a practical process framework for practical applications. 
At the model level, we partitioned the model into four major components and employed three learning methods for supervision. 
We conducted a wide spectrum of experiments to validate their effectiveness.
Moreover, we conducted experiments focusing on the data source to investigate the impact of flight altitude, satellite image scale, and satellite image time on the positioning results.
Last, our proposed baseline model achieved remarkable results on the DenseUAV dataset.
It is worth mentioning that due to the challenges posed by modality differences between UAV and satellite data, as well as the characteristics of UAV self-positioning task, we have observed that the Transformer models outperform CNNs significantly.

However, DenseUAV still has some limitations, primarily related to scene coverage and altitude range. 
The coverage of DenseUAV positioning is only for low-altitude urban scenes. 
No research has been done for other scenarios.
This leaves a space for future work and studies in the UAV self-positioning task.

\bibliographystyle{IEEEtran}
\bibliography{IEEEabrv, DenseUAV}

\end{document}